\documentclass[10pt,twocolumn,letterpaper]{article}

\usepackage{3dv}
\usepackage{times}
\usepackage{graphicx}
\usepackage{amsmath}
\usepackage{amssymb}
\usepackage{enumitem}
\usepackage{subcaption}

\newcommand{\threesixty}{$360^{\circ}$}

\threedvfinalcopy 

\def\threedvPaperID{4} 

\ifthreedvfinal\fi
\begin{document}

\title{Pano Popups: Indoor 3D Reconstruction with a Plane-Aware Network}

\author{Marc Eder\\
Univeristy of North Carolina at Chapel Hill\\
Chapel Hill, NC\\
{\tt\small meder@cs.unc.edu}
\and
Pierre Moulon\\
Zillow Group\\
Seattle, WA\\
{\tt\small pierrem@zillow.com}
\and
Li Guan\\
Wormpex AI Research\\
Seattle, WA\\
{\tt\small li.guan@bianlifeng.com}
}

\maketitle

\begin{abstract}
In this work we present a method to train a plane-aware convolutional neural network for dense depth and surface normal estimation as well as plane boundaries from a single indoor \threesixty image. Using our proposed loss function, our network outperforms existing methods for single-view, indoor, omnidirectional depth estimation and provides an initial benchmark for surface normal prediction from \threesixty images. Our improvements are due to the use of a novel plane-aware loss that leverages principal curvature as an indicator of planar boundaries. We also show that including geodesic coordinate maps as network priors provides a significant boost in surface normal prediction accuracy. Finally, we demonstrate how we can combine our network's outputs to generate high quality 3D ``pop-up" models of indoor scenes.
\end{abstract}

\section{Introduction}
Omnidirectional imaging is currently experiencing a surge in popularity, thanks to the advent of interactive panorama photo sharing on social media platforms, the rise of small, affordable cameras like the Ricoh Theta and Samsung Gear360, and the host of potential applications that arise from capturing wide field of view (FoV) in a single frame. At the same time, deep learning has never been a more useful tool for solving computer vision tasks from object recognition to 3D reconstruction. In order to fully utilize this rising form of media, we must extend existing deep learning methods to the omnidirectional domain. Unfortunately, this is not necessarily a trivial task. 

\begin{figure}[ht]
\begin{center}
\includegraphics[width=1.0\linewidth]{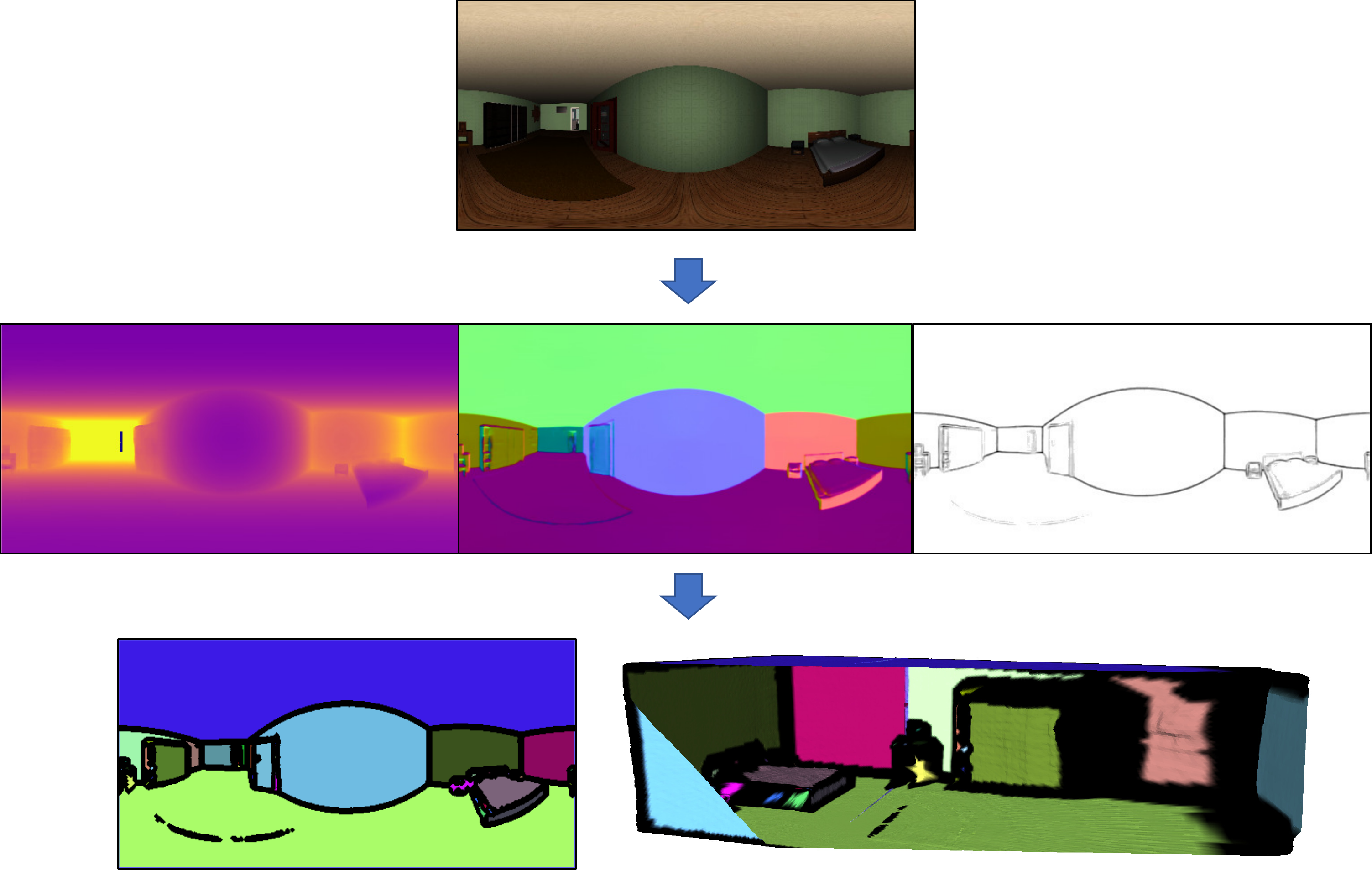}
\end{center}
   \caption{Visualization of our paper's contributions. Given an RGB omnidirectional image (top), we predict depth, surface normals, and plane boundary maps (middle) with state-of-the-art accuracy. Then we show we can use this information to achieve a planar segmentation and 3D reconstruction of the input image (bottom).}
\label{fig:preview}
\end{figure}

Due to the radically different camera models, deep networks trained on perspective images do not transfer well to omnidirectional images. Omnidirectional images replace the concept of the image plane with that of the image sphere. Yet because we require a 2D planar representation of the image, omnidirectional cameras typically provide outputs as $180^{\circ} \times 360^{\circ}$ FoV equirectangular projections. This representation of the spherical image, while compact, suffers from significant horizontal distortion, especially near the poles.

While there have been a number of efforts to handle the difficulties of equirectangular projections  \cite{cohen2017convolutional,cohen2018spherical, coors2018spherenet, eder2019convolutions, su2017learning, tateno2018distortion}, we are interested in exploring their possible uses. There is excitement over the range of applications of omnidirectional imaging from head-mounted displays to medical scopes to autonomous vehicles. In this paper, we target indoor scene modeling.

Perspective image methods are impeded by a small FoV that is more likely to be limited by featureless, homogeneous regions in an indoor scene. With the larger FoV in \threesixty images, these homogeneous regions can be reasoned about in the larger context of the scene. Our goal is to predict the dense depth and surface normals for a piecewise-planar reconstruction of the scene. This objective differs from much of the existing work that uses omnidirectional images for indoor 3D modeling. Those, such as RoomNet \cite{lee2017roomnet} and LayoutNet \cite{zou2018layoutnet}, aim to generate a simple model of the scene by leveraging a Manhattan World constraint to estimate the \textit{dominant} planes. That type of model is useful for determining the shapes of rooms and floor-plans of buildings, but not for modeling the objects that comprise the captured scene. While we, too, are essentially estimating planes in the scene, we aim for a more fine-grained model in order to better capture these important details. To this end, we relax the Manhattan constraint to a simple planar one. That is, we assume only that our scene is piecewise-planar.

We use a convolutional neural network (CNN) to predict depth and surface normal estimates per pixel as well as a map of the plane boundaries in the image. We enforce the planar assumption by using a plane-aware loss function that modifies each pixel's contribution to the learning based on its principal curvature. Using our network outputs, we then generate high quality 3D planar models of the scene as seen in Figure \ref{fig:preview}.

We summarize our contributions in this paper as follows:
\begin{itemize}[topsep=0pt,itemsep=-1ex,partopsep=1ex,parsep=1ex]
    \item We propose a plane-aware cost function to estimate depth, surface normals, and plane boundaries from a single \threesixty image.
    \item We demonstrate that the inclusion of geodesic coordinate maps as extra inputs to the network improves surface normal prediction from omnidirectional images.
    \item We qualitatively show that our network can be used to generate a 3D planar model from a single \threesixty image.
\end{itemize}

\begin{figure*}[ht]
\begin{center}
\includegraphics[width=1.0\linewidth]{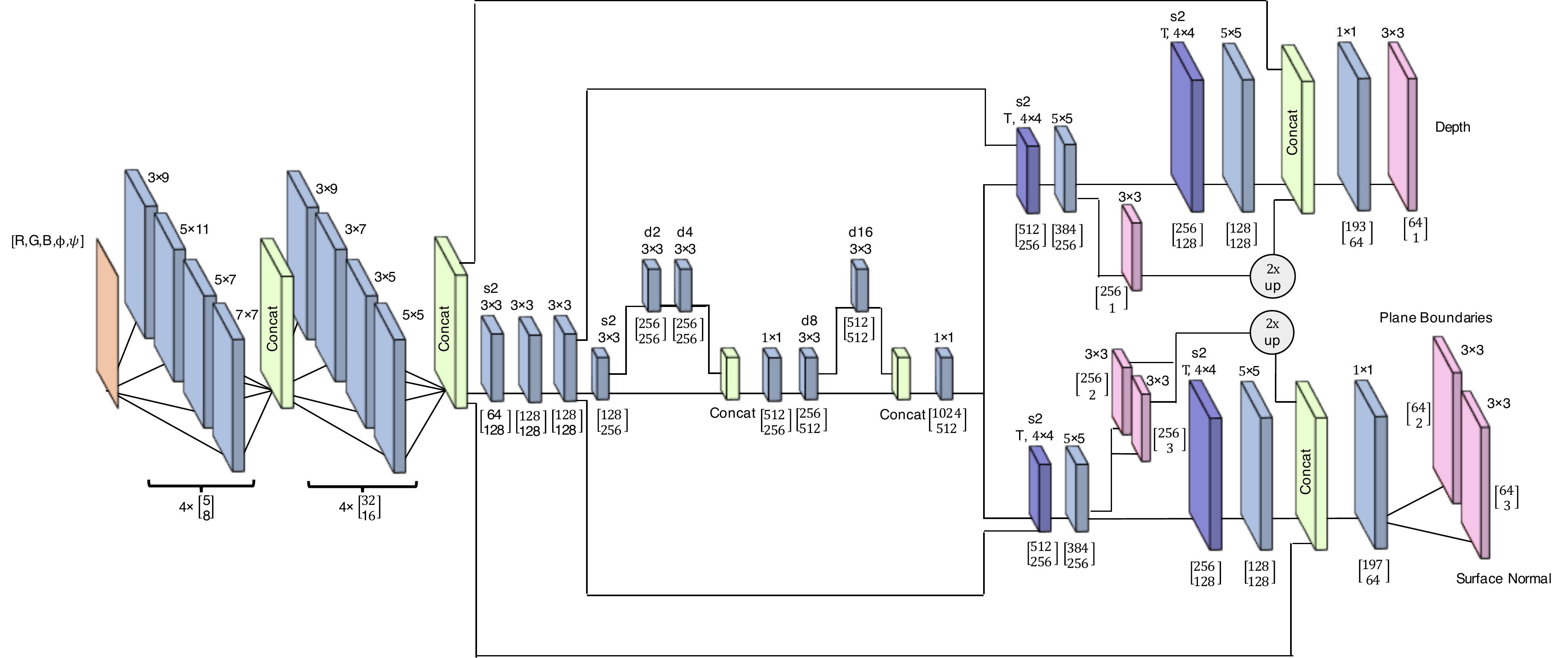}
\end{center}
\vspace{-2mm}
\caption{Overview of our network architecture. The vectors next to each layer are $\left[ \protect\begin{smallmatrix}\text{input features}\\ \text{output features}\protect\end{smallmatrix} \right]$ for each. Above each layer is the kernel size, `T' indicates transposed convolution, `s\#' indicates stride, and `d\#' indicates dilation. The network follows a similar encoder-decoder model to Zioulis \etal \cite{zioulis2018omnidepth}. There are two decoder branches: one for depth, and the other for normals and curvature. The $2\times$ downsampled predictions are upsampled and used for $1\times$ scale prediction in both branches.}
\label{fig:architecture}
\end{figure*}
\section{Related Work}
\subsection{Single-view estimation}
There is a significant body of existing research on the task of monocular depth estimation from perspective images. One of the first papers to report success in this task was from Saxena \etal \cite{saxena2006learning}, who use a Markov Random Field to infer depth from a blend of local and global image features. With the advent of practical deep learning, more recent methods have focused on applying CNNs to estimate depth. Eigen \etal \cite{NIPS2014_5539} present a CNN for depth estimation that uses multi-scale predictions to provide coarse and fine supervision for the depth predictions.  Eigen \etal \cite{eigen2015predicting} built on that work to simultaneously generate surface normal predictions and semantic labels as well. Dharmasiri \etal \cite{dharmasiri2017joint} follow a similar network design but replace semantic label prediction with principal curvature prediction. Our network architecture has some commonalities with the aforementioned, primarily in our use of multi-scale predictions and similar prediction modalities. However, our goal is more aligned with that of Qi \etal \cite{qi2018geonet} who propose a method for enforcing geometric consistency in the network outputs. In that work, the authors use the depth predictions to refine normal predictions and vice versa. In our case, we use a plane-aware loss to make our network predictions geometrically consistent. Our objective is also somewhat similar to that of Liu \etal \cite{liu2018planenet} who predict a planar segmentation of the scene. However, they rely on a separate plane classification branch in their network and are limited to a fixed number of planes. We use a parametric definition of a plane derived from the principal curvature map and are thus unlimited in the number of planes we can predict.

There have been other recent works in monocular depth estimation that, while interesting and useful, are not currently feasible for our task. Godard \etal \cite{godard2017unsupervised} use stereo image pairs to train a model for monocular depth estimation using an image reconstruction loss. In our case, we only have access to monocular images. Li and Snavely \cite{li2018megadepth} train a network on a dataset built from large-scale, unordered image collections. Alas, there is not yet such a repository for omnidirectional images. 

\subsection{Omnidirectional images}
The primary distinction between our work and those presented above is the mode of our input data. Most research in monocular depth estimation has relied thus far on perspective image projections. We instead operate on equirectangular image projections, which image a spherical capture oo a plane. This representation carries high levels of distortion. There is an active branch of research in developing solutions to account for these factors. Su and Grauman \cite{su2017learning} propose a transfer learning approach to train networks to operate on equirectangular projections. Using an existing perspective-projection-trained network as the target, they train an equirectangular network with a learnable adaptive convolutional kernel to match the outputs. Tateno \etal \cite{tateno2018distortion} present a distortion-aware convolutional kernel that convolves over the sampling grid transformed by a distortion function. In this way, the network can be trained on perspective images and still perform effectively on spherical projections. Coors \etal \cite{coors2018spherenet} independently derive the same operation and show that it can be highly effective for object detection on \threesixty images. Both methods train on perspective images and evaluate on spherical projections. Another promising method is the spherical convolution derived by Cohen \etal \cite{cohen2017convolutional} \cite{cohen2018spherical}. Spherical convolutions address the nuances of spherical projections by filtering \textit{rotations} of the feature maps rather than \textit{translations}. Most recently, Eder and Frahm \cite{eder2019convolutions} demonstrate that resampling spherical images to a subdivided icosahedron substantially improves the performance of CNNs trained on spherical data.  In our work we do not directly address the problem of specialized convolutions. Rather, we explore the application of omnidirectional image inference for the task of indoor 3D modeling. Our work is most similar to that of Zioulis \etal \cite{zioulis2018omnidepth} who estimate depth directly from omnidirectional images.

There is also a growing body of work using \threesixty panorama images to generate indoor scene layouts. Xu \etal \cite{xu2017pano2cad} fuse object detection and 3D geometry estimation use Bayesian inference to generate 3D room layouts from a single \threesixty image. Rather than dividing the problem into sub-tasks, Lee \etal \cite{lee2017roomnet} use an end-to-end CNN to generate a 3D room layout from a single perspective image. Zhou \etal \cite{zou2018layoutnet} improve this technique by incorporating vanishing point alignment and prediction additional layout elements to their model. All of the aforementioned layout generation models assume a Manhattan World in their predictions. While this may be useful for common room shapes, it is too simple a prior for general indoor scene modeling. Our work focuses on a more complete indoor 3D model, so we relax this Manhattan constraint to a planar one.

\section{Plane-Aware Estimation}
We present a CNN that estimates dense depth and surface normal predictions as well as a planar boundary map from a single \threesixty image. To learn depth and normal prediction, we supervise training with ground truth values. Observing that a non-zero principal curvature indicates the presence of a planar boundary, we supervise training for the planar boundary map using the $L_2$ norm of the principal curvature.

\subsection{Network architectures}
We analyze our plane-aware loss function using a network based on the RectNet architecture used by Zioulis \etal \cite{zou2018layoutnet}. Our network uses the same encoder-decoder structure with rectangular filter banks on the input layers, but with two decoder branches: one for depth predictions and one for joint surface normal and plane boundary map prediction. We also include skip connections from encoder to decoder layers as in U-Net from Ronneberger \etal \cite{ronneberger2015u}, as we observe it improves performance. Our network takes a five-channel input: an RGB equirectangular projection and the associated geodesic map containing latitude and longitude coordinates for each pixel. This design is based on the observation that distortion in equirectangular projections is location dependent. Given that these images are indexed by their geodesic coordinates, given in latitude and longitude, we provide the network with location information in the form of a geodesic coordinate map of the image. We find that this provides a significant boost in performance for surface normal prediction in particular and discuss it in more detail in Section \ref{sec:coordprior}. Figure \ref{fig:architecture} provides a detailed overview of our network.
\begin{figure*}
    \centering
    \setlength{\fboxsep}{0pt}
    \setlength{\fboxrule}{0.5pt}
    \begin{subfigure}[b]{0.15\textwidth}
        \centering
        \fbox{\includegraphics[width=\textwidth]{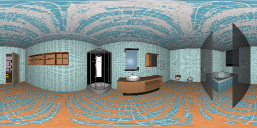}}
    \end{subfigure}
    ~
    \centering
    \begin{subfigure}[b]{0.15\textwidth}
        \centering
        \fbox{\includegraphics[width=\textwidth]{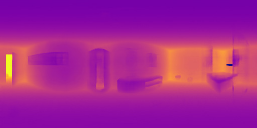}}
    \end{subfigure}
    ~
    \centering
    \begin{subfigure}[b]{0.15\textwidth}
        \centering
        \fbox{\includegraphics[width=\textwidth]{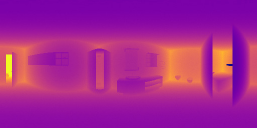}}
    \end{subfigure}
    ~
    \centering
    \begin{subfigure}[b]{0.15\textwidth}
        \centering
        \fbox{\includegraphics[width=\textwidth]{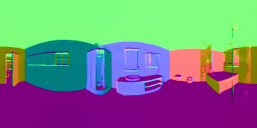}}
    \end{subfigure}
    ~
    \centering
    \begin{subfigure}[b]{0.15\textwidth}
        \centering
        \fbox{\includegraphics[width=\textwidth]{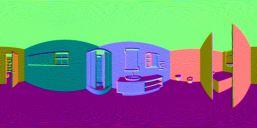}}
    \end{subfigure}
    ~
    \centering
    \begin{subfigure}[b]{0.15\textwidth}
        \centering
        \fbox{\includegraphics[width=\textwidth]{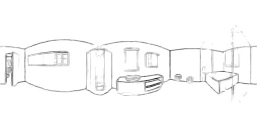}}
    \end{subfigure}\\
\vspace{2pt}

\begin{subfigure}[b]{0.15\textwidth}
        \centering
        \fbox{\includegraphics[width=\textwidth]{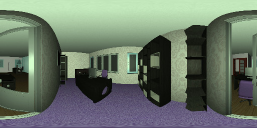}}
    \end{subfigure}
    ~
    \centering
    \begin{subfigure}[b]{0.15\textwidth}
        \centering
        \fbox{\includegraphics[width=\textwidth]{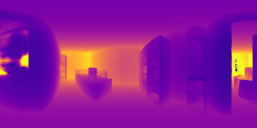}}
    \end{subfigure}
    ~
    \centering
    \begin{subfigure}[b]{0.15\textwidth}
        \centering
        \fbox{\includegraphics[width=\textwidth]{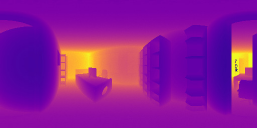}}
    \end{subfigure}
    ~
    \centering
    \begin{subfigure}[b]{0.15\textwidth}
        \centering
        \fbox{\includegraphics[width=\textwidth]{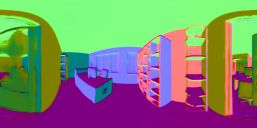}}
    \end{subfigure}
    ~
    \centering
    \begin{subfigure}[b]{0.15\textwidth}
        \centering
        \fbox{\includegraphics[width=\textwidth]{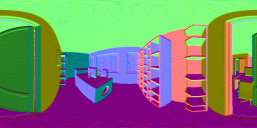}}
    \end{subfigure}
    ~
    \centering
    \begin{subfigure}[b]{0.15\textwidth}
        \centering
        \fbox{\includegraphics[width=\textwidth]{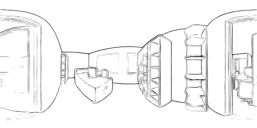}}
    \end{subfigure}\\
\vspace{2pt}
\begin{subfigure}[b]{0.15\textwidth}
        \centering
        \fbox{\includegraphics[width=\textwidth]{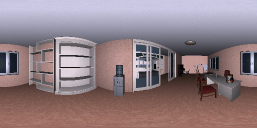}}
        \caption{RGB Input}
    \end{subfigure}
    ~
    \centering
    \begin{subfigure}[b]{0.15\textwidth}
        \centering
        \fbox{\includegraphics[width=\textwidth]{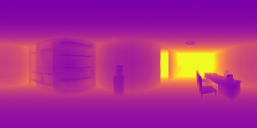}}
        \caption{Pred. Depth}
    \end{subfigure}
    ~
    \centering
    \begin{subfigure}[b]{0.15\textwidth}
        \centering
        \fbox{\includegraphics[width=\textwidth]{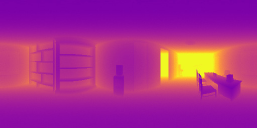}}
        \caption{GT Depth}
    \end{subfigure}
    ~
    \centering
    \begin{subfigure}[b]{0.15\textwidth}
        \centering
        \fbox{\includegraphics[width=\textwidth]{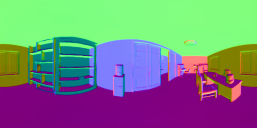}}
        \caption{Pred. Normals}
    \end{subfigure}
    ~
    \centering
    \begin{subfigure}[b]{0.15\textwidth}
        \centering
        \fbox{\includegraphics[width=\textwidth]{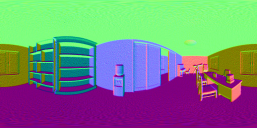}}
        \caption{GT Normals}
    \end{subfigure}
    ~
    \centering
    \begin{subfigure}[b]{0.15\textwidth}
        \centering
        \fbox{\includegraphics[width=\textwidth]{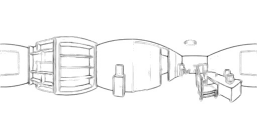}}
        \caption{Pred. Boundaries}
    \end{subfigure}\\
    \label{fig:omnidepthresults}
\caption{Examples of our network predictions on the SUMO dataset \cite{sumo}. Observe that the plane boundary maps only include the geometric edges in the scene. For example, they do not include the highly textured floor and ceiling in the top row input.}
\end{figure*}

\subsection{Training}
Recall our premise that each scene is piecewise-planar. This assumption provides a few constraints. First, each scene should be segmented by some web of edges that define the boundaries between each plane. Second, each planar region should have a constant depth gradient and all pixels within should have the same surface normal. Furthermore, the principal curvature, which is effectively the second derivative of depth, should be zero. Lastly, the depth and normal predictions within a planar region should satisfy the plane equation $n^TX + d = 0$, where $n$ is the normal, $X$ is the 3D point, and $d$ is the plane's distance from the origin.

We enforce these constraints through a multi-scale, multi-task loss function. We compute individual losses over the depth, surface normals, and plane boundary map predictions as well as a loss over the plane distance prediction for each pixel, denoted as $L_{z}$, $L_{n}$, $L_{C}$, and $L_{D}$, respectively. This last term is computed as a function of both the depth and normal predictions, which encourages planar consistency. Each of the losses is also weighted using a plane-aware function $P(x)$. For the depth, curvature, and plane distance losses, we use the reverse Huber, or BerHu, loss proposed by Laina \etal \cite{laina2016deeper}. This loss is given as
\begin{equation}
    B(x) = 
    \begin{cases}
        |x| & |x| \leq T \\
        \frac{x^2 + T^2}{2T} & |x| > T
    \end{cases}
\label{eq:berhu}
\end{equation}

where we adjust $T$ on a per-batch basis to be 20\% of the max per-batch error as in \cite{laina2016deeper}. Our plane-aware function $P(x)$ weights the impact of each pixel $x$ to the loss by the $L_2$ norm of its ground truth principal curvature, $c^*$:
\begin{equation}\label{eq:penalty}
    P(x) = x e^{-||c^*||}
\end{equation}

As curvature is zero on a planar surface, this term gives full weight to all pixels that lie on planes. However, pixels that fall along sharp plane boundaries and thus have higher curvatures will have their contribution to the loss down-weighted. This is similar to the texture-edge-aware loss weighting used by Godard \etal \cite{godard2017unsupervised}, except that we use the curvature values instead of intensity gradients. Our formulation makes more sense for our task, given that we are interested in planar boundaries rather than texture ones.

Each component of the loss is given below. The subscript $i$ denotes the $i$-th pixel in the image; $z_i$ is depth, $n_i$ is normal, and $c_i$ is curvature.
\begin{equation}\label{eq:depthloss}
    L_{z} = \frac{1}{|\mathcal{M}|} 
    \sum_{i \in \mathcal{M}} 
    P\left(B\left(z_{i} - z_{i}^{*}\right)\right)
\end{equation}
\begin{equation}\label{eq:normalloss}
    L_{n} = \frac{1}{|\mathcal{M}|} 
    \sum_{i \in \mathcal{M}} 
    P\left(-{n_i}^T n_i^* \right)
\end{equation}
\begin{equation}\label{eq:curvloss}
    L_{C} = \frac{1}{|\mathcal{M}|} 
    \sum_{i \in \mathcal{M}} 
    P\left(B\left(||c_{i}||_2 - ||c_{i}^{*}||_2\right) + \eta||c_i||_1\right)
\end{equation}
\begin{equation}\label{eq:planeloss}
    L_{D} = \frac{1}{|\mathcal{M}|} 
    \sum_{i \in \mathcal{M}} 
    P\left(B\left(n_i^T X_{z_i} - {n_i^*}^T X_{z_i^*}\right)\right)
\end{equation}
where $\mathcal{M}$ is the relevant output map and the asterisks denote ground truth values. In Equation (\ref{eq:planeloss}), $X_{z_i} = z_i \hat{b}$ where $\hat{b}$ is the directional unit vector from the camera center to pixel $i$ on the sphere, i.e. $X_{z_i}$ is the back-projected 3D point.

It is worth noting that other single-view depth estimation papers typically include an $L_2$ penalty on the gradient of the depth or disparity prediction to account for homogeneous regions where depth may be ambiguous \cite{garg2016unsupervised, zioulis2018omnidepth}. However, this term is known to lead to over-smoothing, especially for surfaces that are not fronto-planar to the camera. In the case of \threesixty images, where depth is defined as the distance from a 3D point to the camera center (rather than to the image plane), this gradient penalty would encourage the prediction of a circular scene wherein each point is locally fronto-planar to the camera. Thus, we do not penalize the depth gradient at all. In the planar boundary map prediction, however, we do include an $L_1$ penalty to encourage sparsity in the edge predictions.

Our total loss is thus the sum of all of these terms at two scales weighted by some hyper-parameters $\alpha$, $\beta$, $\gamma$, and $\zeta$:
\begin{equation}
    L_{total} = \sum_{s\in\{0,1\}} \alpha_s L_{z} + \beta_s L_{\hat{n}} + \gamma_s L_{C} + \zeta_s L_{D}
\label{eq:totalloss}
\end{equation}

We empirically set the hyper-parameters to balance the contribution of each component loss. In our reported results, $\alpha_{0} = 0.3$, $\alpha_{1} = 0.6$,  $\beta_{0} = 0.1$, $\beta_{1} = 0.4$,  $\gamma_{0} = 0$, $\gamma_{1} = 0.3$,  $\zeta_{0} = 0.3$, and $\zeta_{1} = 0.6$. The $L_1$ penalty coefficient in Equation (\ref{eq:curvloss}) is always $\eta = 0.1$. Nonetheless, we observed that small changes to these hyper-parameters have negligible effects on the network training. Note that we do not use any loss for planar boundary map prediction for the down-scaled prediction ($\gamma_{0}$) as we observed that it made no impact in the final plane boundary map.  We train the network for $20$ epochs with a batch size of $10$ and use the Adam optimizer \cite{kingma2014adam} with an initial learning rate of $2e^{-4}$ decayed by half every $3$ epochs.
\begin{table*}[ht]
\small
\begin{center}
\begin{tabular}{|c|c|c|c|c|c|c|c|c|}
\hline
Loss & AbsRel $\downarrow$ & SqRel $\downarrow$ & RMSLin $\downarrow$ & RMSLog $\downarrow$ & $\delta<1.25 \uparrow$ & $\delta<1.25^2 \uparrow$ & $\delta<1.25^3 \uparrow$ \\
\hline\hline
L2 + smoothing \cite{zioulis2018omnidepth} & $0.0447$ & $0.0191$ & $0.0959$ & $0.2117$ & $0.9721$ & $0.9924$ & $0.9971$ \\
\textbf{Plane-aware (ours)} & $\mathbf{0.0275}$ & $\mathbf{0.0113}$ & $\mathbf{0.0715}$ & $\mathbf{0.1658}$ & $\mathbf{0.9851}$ & $\mathbf{0.9953}$ & $\mathbf{0.9980}$ \\
\hline\hline
Ablation & \multicolumn{7}{c|}{} \\
\hline
L2 instead of BerHu & $0.0362$ & $0.0150$ & $0.0863$ & $0.1907$ & $0.9789$ & $0.9940$ & $0.9976$ \\
No curvature penalty & $0.0294$ & $0.0122$ & $0.0743$ & $0.1720$ & $0.9847$ & $0.9949$ & $0.9979$ \\
No plane loss & $0.0286$ & $0.0122$ & $0.0742$ & $0.1706$ & $0.9839$ & $0.9948$ & $0.9976$ \\
\hline
\end{tabular}
\end{center}
\vspace{-4mm}
\caption{Depth estimation results comparing out loss to alternatives and ablated forms. Our baseline, L2 + smoothing, is the approach taken by Zioulis \etal \cite{zioulis2018omnidepth}. We also evaluate using an L2 loss in place of BerHu, training the network without the curvature-aware penalty, Eq. (\ref{eq:penalty}), removing the planar-consistency regularizer, Eq. (\ref{eq:planeloss}), and omitting the joint plane boundary map prediction, Eq. (\ref{eq:curvloss}).}
\vspace{-2mm}
\label{tab:depthresults}
\end{table*}

\section{Evaluation}
In this section we evaluate our proposed plane-aware depth and normal estimation. First, we demonstrate the benefit of our plane-aware loss through comparison to a baseline, the loss used by Zioulis \etal \cite{zioulis2018omnidepth}, as well as in a series of ablation experiments. Second, we demonstrate the importance of predicting surface normals rather than relying on derived normals from predicted depth. We then examine the effect of including coordinate priors as inputs to the network. Finally, we qualitatively show how we can leverage the predicted plane boundary map to create 3D reconstructions in Section \ref{sec:planes}.

\subsection{Dataset}
We train and evaluate our method using the Scene Understanding and Modeling (SUMO) dataset \cite{sumo}, a collection of 58,631 computer generated omnidirectional images of indoor scenes derived from SunCG \cite{suncg}. As released, the SUMO dataset contains RGB-D cube map images with a cube face dimension of $1024$ pixels. To prepare this data for our experiments, we resample the cube maps to $256 \times 512$ pixel equirectangular images using bilinear interpolation for color information and nearest-neighbor interpolation for depth. For the purposes of surface normal and principal curvature prediction, we augment the dataset with normal and curvature maps for each image as well. We derive the ground truth normal maps from the provided images by first resampling them to the vertices of icosahedral triangular mesh as in \cite{eder2019convolutions}, scaling each vertex by the ground truth depth, computing the surface normal for each face, and rendering the normal maps back into an equirectangular projection. For the ground truth planar boundary maps, we use the $L_2$ norm of the principal curvature. The curvature maps are derived as in \cite{spek2017fast} using the eigenvalues of the $2\times2$ matrix given by:

$$\mathbf{II} = 
\left[\begin{array}{cc}
    du & dv 
\end{array}\right]
\left[\begin{array}{cc}
    A & B \\ 
    B & C  
\end{array}\right]
\left[\begin{array}{c}
    du \\
    dv
\end{array}\right]
$$

where $u$ and $v$ are vectors that, with the surface normal $\hat{n}$, form an orthonormal basis at a given point $(x,y)$. $A$, $B$, and $C$ are defined by the derivatives of the the surface normal at that point:

$$ A = -\frac{\delta \hat{n}(x,y)}{dx} \cdot u,  B = -\frac{\delta \hat{n}(x,y)}{dy} \cdot u,  C = -\frac{\delta \hat{n}(x,y)}{dx} \cdot v$$

\begin{table*}[ht]
\small
\begin{center}
\begin{tabular}{|c|c|c|c|c|c|}
\hline
Method & Avg. Ang. $\downarrow$ & $<7.5^{\circ} \uparrow$ & $<15^{\circ} \uparrow$ & $<30^{\circ} \uparrow$ & $<45^{\circ} \uparrow$ \\
\hline\hline
\textbf{Plane-aware + Lat./Lon.} & $\mathbf{6.5630^{\circ}}$ & $\mathbf{0.7683}$ & $\mathbf{0.9358}$ & $\mathbf{0.9728}$ & $\mathbf{0.9855}$ \\
\hline
Derived from depth & $14.3272^{\circ}$ & $0.4675$ & $0.7229$ & $0.8796$ & $0.9351$ \\
No curvature penalty & $6.6645^{\circ}$ & $0.7643$ & $0.9330$ & $0.9715$ & $0.9849$ \\
No plane loss & $6.6318^{\circ}$ & $0.7647$ & $0.9324$ & $0.9720$ & $\mathbf{0.9855}$ \\
\hline
No coordinates & $6.9070^{\circ}$ & $0.7608$ & $0.9288$ & $0.9680$ & $0.9820$ \\
Lat. only & $6.9529^{\circ}$ & $0.7599$ & $0.9280$ & $0.9674$ & $0.9815$ \\
Lon. only & $6.5833^{\circ}$ & $0.7676$ & $0.9353$ & $0.9725$ & $0.9853$ \\
\hline
\end{tabular}
\end{center}
\vspace{-3mm}
\caption{Surface normal prediction results. Due to a dearth of existing methods for surface normal prediction on omnidirectional images, we evaluate against surface normals derived from  predicted depth and perform ablation studies.}
\vspace{-4mm}
\label{tab:normalresults}
\end{table*}

\subsection{Depth estimation}
We evaluate the depth estimation task using the standard set of metrics defined in Eigen \etal \cite{NIPS2014_5539}, shown in Table \ref{tab:depthresults}. Because depth estimates are subject to the arbitrary scale of the training distribution, we use the median scaling technique given by \cite{zioulis2018omnidepth} to normalize the depth distributions during evaluation. The numbers we report are based on pixels whose ground truth depth falls within the range $\left[ 0, T_{depth} \right]$. We set $T_{depth}$ to be 4.375 standard deviations above the mean of the training set, deriving this value from an analysis of the evaluation threshold used by Zioulis \etal \cite{zioulis2018omnidepth}. To evaluation our proposed loss, we compare to network training under the loss used by Zioulis \etal \cite{zioulis2018omnidepth} as a baseline. This loss is simply an $L2$ minimization with a gradient penalty at two scales, as given by Equation (\ref{eq:omnidepthloss}):
\begin{equation}\label{eq:omnidepthloss}
    L_{baseline} = \frac{1}{|\mathcal{M}|}\sum_{s \in \{0,1\}}\sum_{i \in M}\alpha_s||z_ - z_i^*||_2^2 + \beta_s||\nabla z_||_2^2
\end{equation}

\begin{figure*}
    \centering
    \setlength{\fboxsep}{0pt}
    \setlength{\fboxrule}{0.5pt}

    \raisebox{23pt}{\parbox[b]{.03\textwidth}{1)}}%
    \begin{subfigure}[b]{0.23\textwidth}
        \centering
        \fbox{\includegraphics[width=\textwidth]{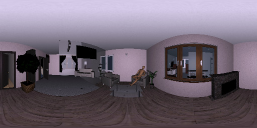}}
    \end{subfigure}
    ~
    \centering
    \begin{subfigure}[b]{0.23\textwidth}
        \centering
        \fbox{\includegraphics[width=\textwidth]{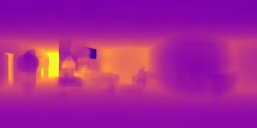}}
    \end{subfigure}
    ~
    \centering
    \begin{subfigure}[b]{0.23\textwidth}
        \centering
        \fbox{\includegraphics[width=\textwidth]{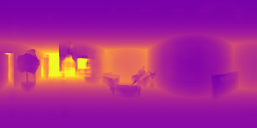}}
    \end{subfigure}
    ~
    \centering
    \begin{subfigure}[b]{0.23\textwidth}
        \centering
        \fbox{\includegraphics[width=\textwidth]{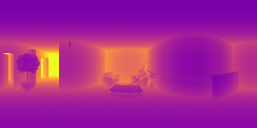}}
    \end{subfigure}\\
\vspace{2pt}
    
    \raisebox{23pt}{\parbox[b]{.03\textwidth}{2)}}%
    \begin{subfigure}[b]{0.23\textwidth}
        \centering
        \fbox{\includegraphics[width=\textwidth]{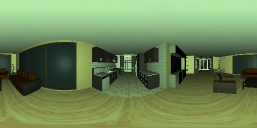}}
    \end{subfigure}
    ~
    \centering
    \begin{subfigure}[b]{0.23\textwidth}
        \centering
        \fbox{\includegraphics[width=\textwidth]{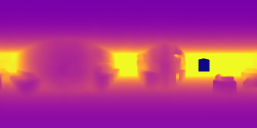}}
    \end{subfigure}
    ~
    \centering
    \begin{subfigure}[b]{0.23\textwidth}
        \centering
        \fbox{\includegraphics[width=\textwidth]{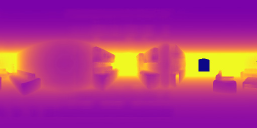}}
    \end{subfigure}
    ~
    \centering
    \begin{subfigure}[b]{0.23\textwidth}
        \centering
        \fbox{\includegraphics[width=\textwidth]{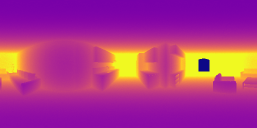}}
    \end{subfigure}\\
\vspace{2pt}    

    \raisebox{40pt}{\parbox[b]{.03\textwidth}{3)}}%
    \begin{subfigure}[b]{0.23\textwidth}
        \centering
        \fbox{\includegraphics[width=\textwidth]{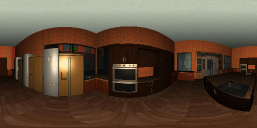}}
        \caption{RGB Input}
    \end{subfigure}
    ~
    \centering
    \begin{subfigure}[b]{0.23\textwidth}
        \centering
        \fbox{\includegraphics[width=\textwidth]{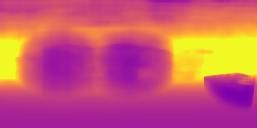}}
        \caption{Baseline}
    \end{subfigure}
    ~
    \centering
    \begin{subfigure}[b]{0.23\textwidth}
        \centering
        \fbox{\includegraphics[width=\textwidth]{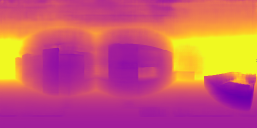}}
        \caption{Ours}
    \end{subfigure}
    ~
    \centering
    \begin{subfigure}[b]{0.23\textwidth}
        \centering
        \fbox{\includegraphics[width=\textwidth]{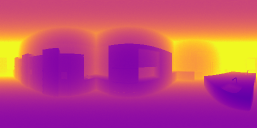}}
        \caption{Ground Truth}
    \end{subfigure}\\
\vspace{-2mm}
\caption{A qualitative comparison of depth predictions using our plane-aware loss compared to the baseline method based on Zioulis \etal \cite{zioulis2018omnidepth}. Notice that our depth estimates are able to capture finer details of the scene.}
\label{fig:qualitativedepth}
\end{figure*}

The results in Table \ref{tab:depthresults} show that our loss formulation outperforms the baseline. We note that the training on synthetic images leads to a high performance for the baseline as well, so we also look to a qualitative analysis to reinforce the effect of our plane-aware formulation. Figure \ref{fig:qualitativedepth} shows a selection of network outputs comparing our loss to the baseline. Observe the finer-grained depth estimate of lounge chair in the center of row (1) and the shelving and counters in rows (2) and (3). We find that training with our proposed plane-aware loss results in sharper details in the resulting depth maps. We posit that this effect is due to extra supervision provided by the ground truth curvature penalty, which limits smoothing on geometric edges.

We perform an ablation study on elements of our loss function, also listed in Table \ref{tab:depthresults}. Among other things, these results demonstrate that our improvement is not simply due to the use of the BerHu loss. We see a moderate impact from both the planar-consistency regularizer as well as the curvature penalty. Interestingly, we found that removing the associated curvature prediction task altogether neither affected the depth or normal prediction accuracy. However, we keep it in the network as it plays a key role in generating the 3D reconstructions, discussed in Section \ref{sec:planes}.

\subsection{Surface normal estimation}
\begin{figure*}
    \centering
    \setlength{\fboxsep}{0pt}
    \setlength{\fboxrule}{0.5pt}

    \raisebox{23pt}{\parbox[b]{.03\textwidth}{1)}}%
    \begin{subfigure}[b]{0.23\textwidth}
        \centering
        \fbox{\includegraphics[width=\textwidth]{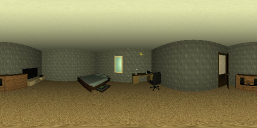}}
    \end{subfigure}
    ~
    \centering
    \begin{subfigure}[b]{0.23\textwidth}
        \centering
        \fbox{\includegraphics[width=\textwidth]{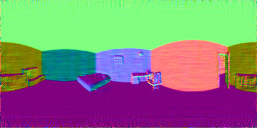}}
    \end{subfigure}
    ~
    \centering
    \begin{subfigure}[b]{0.23\textwidth}
        \centering
        \fbox{\includegraphics[width=\textwidth]{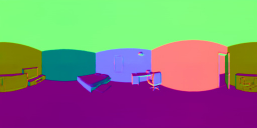}}
    \end{subfigure}
    ~
    \centering
    \begin{subfigure}[b]{0.23\textwidth}
        \centering
        \fbox{\includegraphics[width=\textwidth]{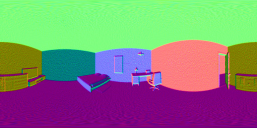}}
    \end{subfigure}\\
\vspace{2pt}
    
    \raisebox{23pt}{\parbox[b]{.03\textwidth}{2)}}%
    \begin{subfigure}[b]{0.23\textwidth}
        \centering
        \fbox{\includegraphics[width=\textwidth]{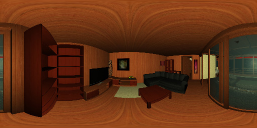}}
    \end{subfigure}
    ~
    \centering
    \begin{subfigure}[b]{0.23\textwidth}
        \centering
        \fbox{\includegraphics[width=\textwidth]{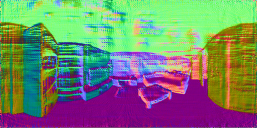}}
    \end{subfigure}
    ~
    \centering
    \begin{subfigure}[b]{0.23\textwidth}
        \centering
        \fbox{\includegraphics[width=\textwidth]{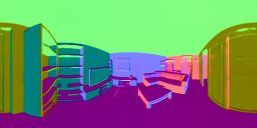}}
    \end{subfigure}
    ~
    \centering
    \begin{subfigure}[b]{0.23\textwidth}
        \centering
        \fbox{\includegraphics[width=\textwidth]{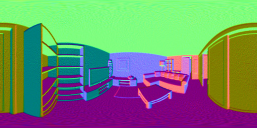}}
    \end{subfigure}\\
\vspace{2pt}    

    \raisebox{40pt}{\parbox[b]{.03\textwidth}{3)}}%
    \begin{subfigure}[b]{0.23\textwidth}
        \centering
        \fbox{\includegraphics[width=\textwidth]{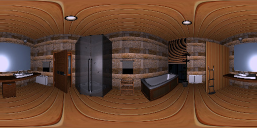}}
        \caption{RGB Input}
    \end{subfigure}
    ~
    \centering
    \begin{subfigure}[b]{0.23\textwidth}
        \centering
        \fbox{\includegraphics[width=\textwidth]{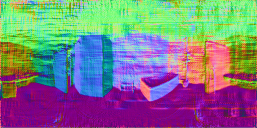}}
        \caption{Derived}
    \end{subfigure}
    ~
    \centering
    \begin{subfigure}[b]{0.23\textwidth}
        \centering
        \fbox{\includegraphics[width=\textwidth]{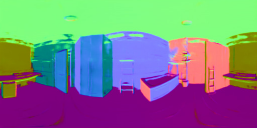}}
        \caption{Predicted}
    \end{subfigure}
    ~
    \centering
    \begin{subfigure}[b]{0.23\textwidth}
        \centering
        \fbox{\includegraphics[width=\textwidth]{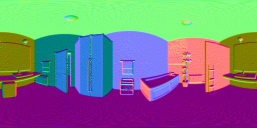}}
        \caption{Ground Truth}
    \end{subfigure}\\
\vspace{-2mm}
\caption{Comparison between our surface normal predictions and those derived the from depth predictions alone. Normal predictions are more reliable than normals derived from depth, as there is no direct dependence between the two predictions. This is important for generating a realistic 3D reconstruction.}
\label{fig:normalresults}
\end{figure*}

For surface normal estimates, we examine pixels that fall within the same valid ground truth depth range. We evaluate the average angular error per pixel as well as the percentage of pixels whose angular error falls within a threshold of the ground truth. Table \ref{tab:normalresults} shows that our loss formulation is useful for improving surface normal prediction. As a baseline we use the surface normals derived from the depth predictions. These results indicate that derived normals are no replacement for an independent surface normal prediction. Our predicted normals are much less susceptible to noisy depth values than their derived counterparts. Figure \ref{fig:normalresults} shows a qualitative comparison of our predicted results compared to the derived normals. When the depth estimation is fairly accurate, the derived normals are only slightly noisier than the prediction, as in row (1). However, in cases where the depth predictions are not as high quality, the predicted normals are often still very good, while the derived normals degrade significantly, as in rows (2) and (3). This effect is why we rely on the indepdendent surface normal prediction branch when generating a 3D reconstruction.
\vspace{-1mm}

\subsection{Geodesic map inputs}\label{sec:coordprior}
\begin{figure}[ht]
\begin{center}
\includegraphics[width=1.0\linewidth]{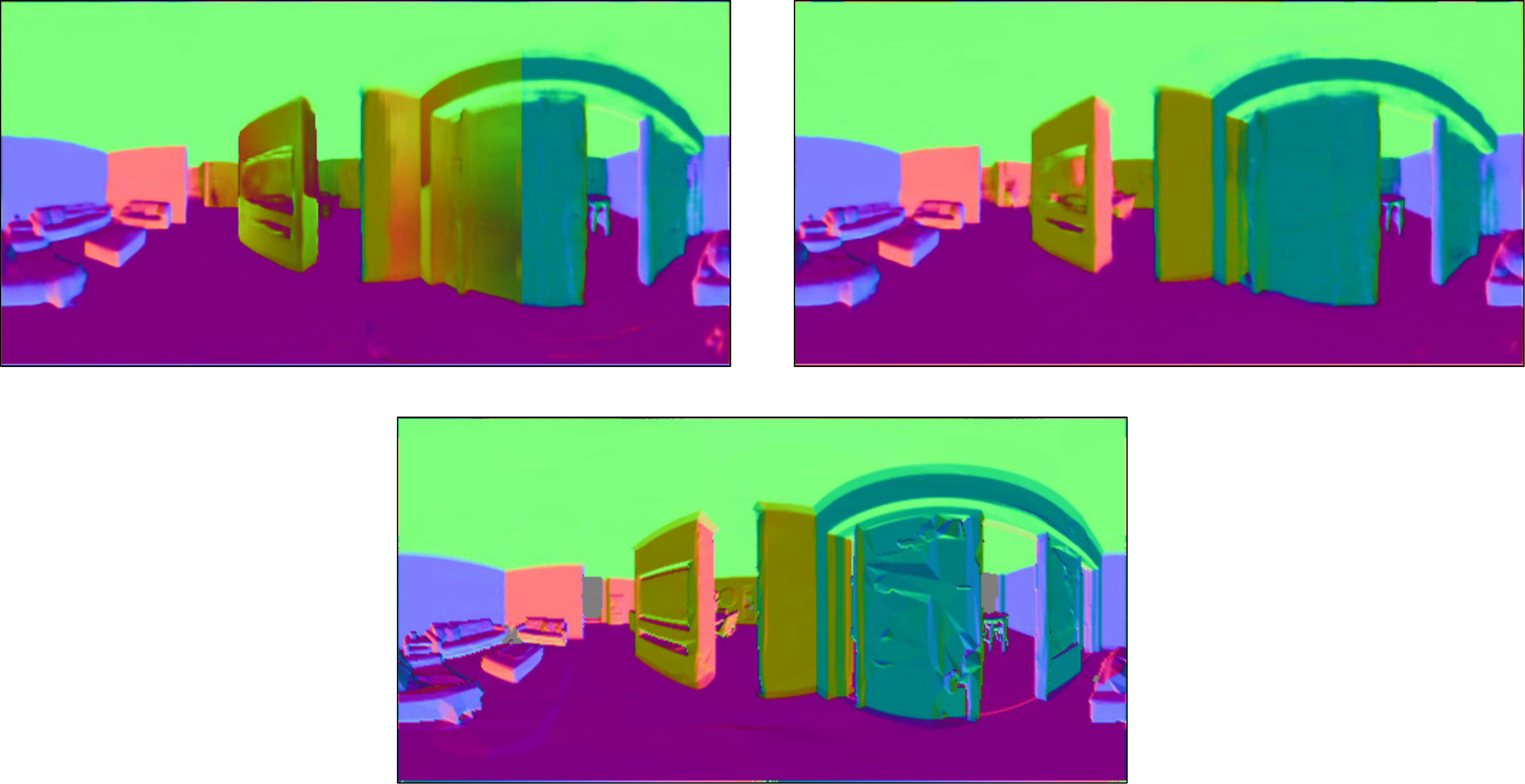}
\end{center}
\vspace{-2mm}
\caption{Demonstrating the impact of geodesic map inputs on surface normal prediction. TL: output without the geodesic maps, TR: output with geodesic maps, B: ground truth. Notice the error in the large wall in the center of the image.}
\label{fig:geodesic}
\end{figure}

We also delve deeper into the impact of the latitude and longitude map priors in the network. Fixing all other aspects of the network, we evaluate the performance of our network on the SUMO dataset with and without the geodesic map channels. Consistent with our expectations, the results in the bottom block of Table \ref{tab:normalresults} suggest that the geodesic map inputs have a positive impact in surface normal estimation. We surmise that the geodesic map helps the network disambiguate the orientation of the surface normal. It is notable that without the geodesic map, we see errors occur at the peak point of barreling on planes in the equirectangular projection as in the top-left image in Figure \ref{fig:geodesic}. Interestingly, longitude provides the most important information, which aligns with what we observe in Figure \ref{fig:geodesic}: predictions changing abruptly along the rows.

Because the equirectangular grid is indexed by spherical coordinates rather than a Cartesian grid, the distance between adjacent pixels is row-dependent as well. Adjacent pixels nearer to the top and bottom of the image actually lie closer together on the sphere than adjacent pixels near the middle of the image do. This sampling scheme is problematic for CNNs because the convolution operation's translation equivariance inherently assumes an even sampling. Somehow the network needs to learn to map the geodesic sampling to a Cartesian one. Our experiments suggest that including the geodesic maps as extra input channels is a useful way to pass this information to the network. These findings line up with the results of Liu \etal \cite{liu2018intriguing} who show that incorporating pixel location information can help a network learn some degree of translation dependence, which is what we also need to achieve. 

\begin{figure}[ht]
\begin{center}
\includegraphics[width=0.8\linewidth]{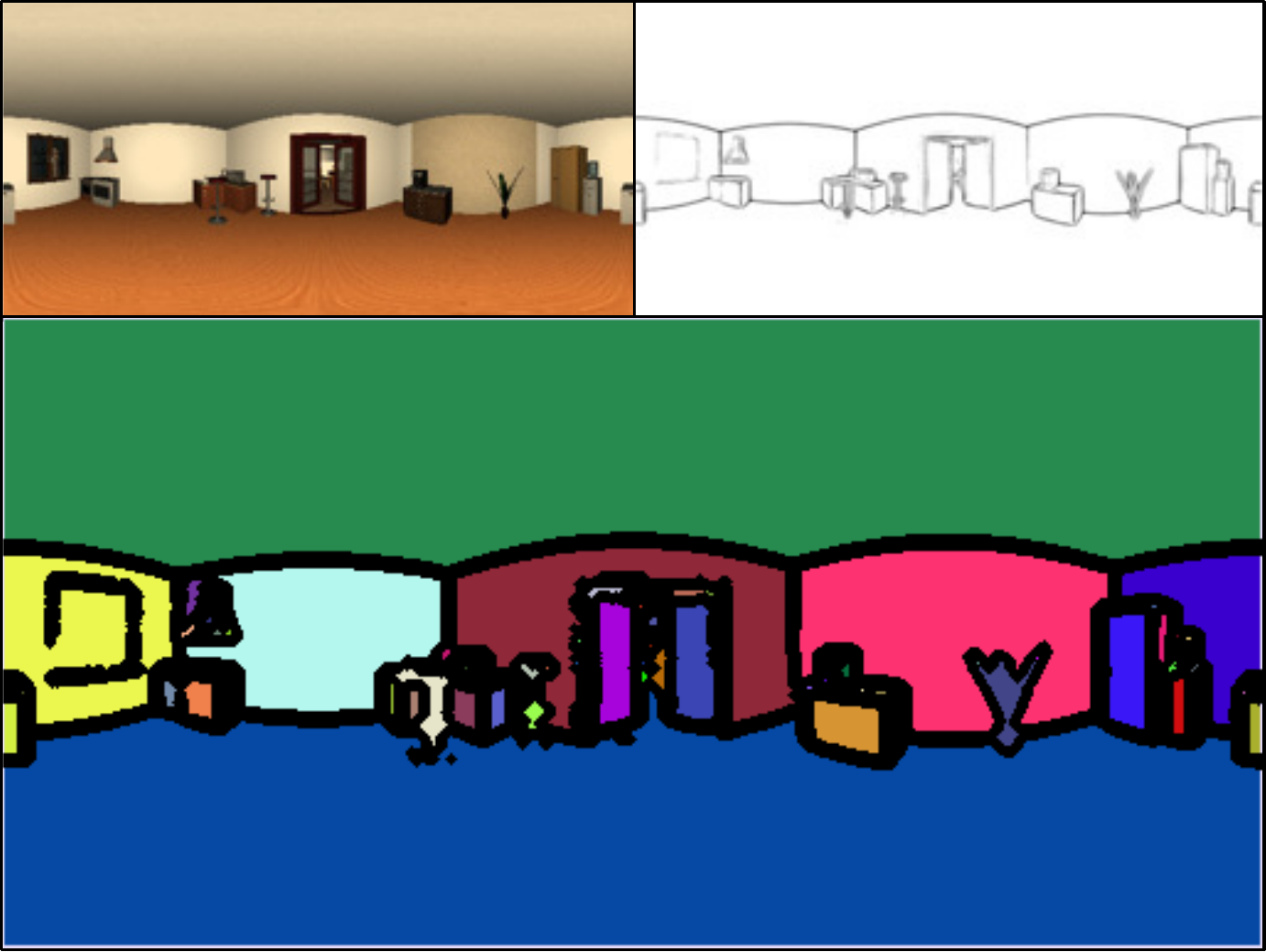}
\end{center}
\vspace{-2mm}
\caption{Plane segmentation using the plane prediction output from our network. TL: RGB input, TR: plane boundary prediction, B: plane boundary segmentation color-coded by label.}
\vspace{-2mm}
\label{fig:planeseg}
\end{figure}
\section{3D Planar Model Reconstruction} \label{sec:planes}
An important consequence of our planarity assumption is that the network provides all of the information necessary to detect and segment planes in the input images. By defining these planes, we can generate ``pop-up" models from a single image, as proposed by Hoiem \etal \cite{hoiem2005automatic}. Indoor omnidirectional images are uniquely suited to produce these types of reconstructions as they are capable of capturing entire rooms in a single image.

\begin{figure}[ht]
\begin{center}
\includegraphics[width=0.48\linewidth]{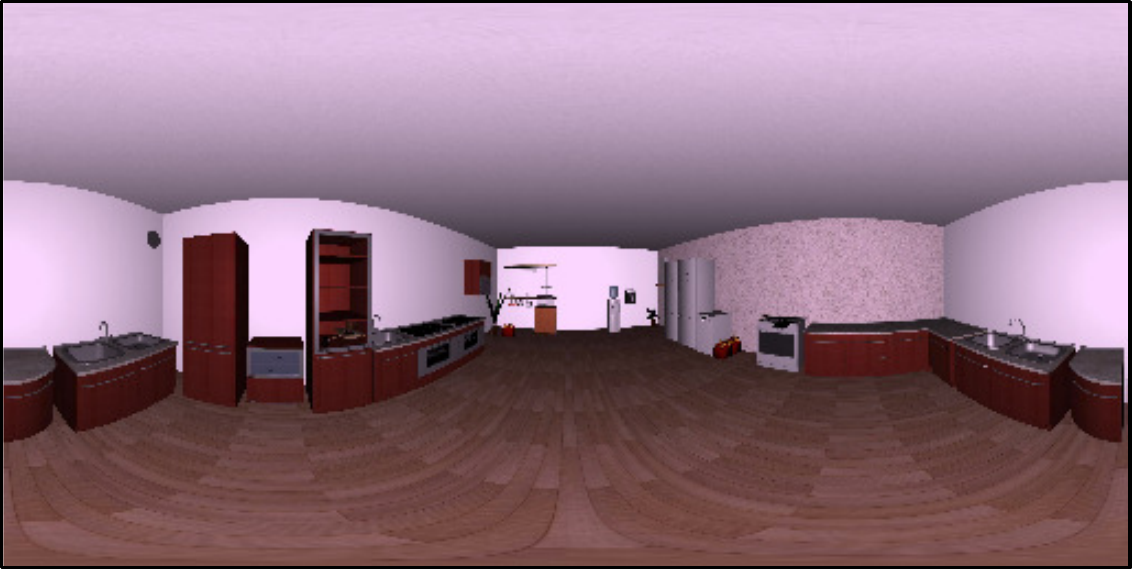}
\includegraphics[width=0.48\linewidth]{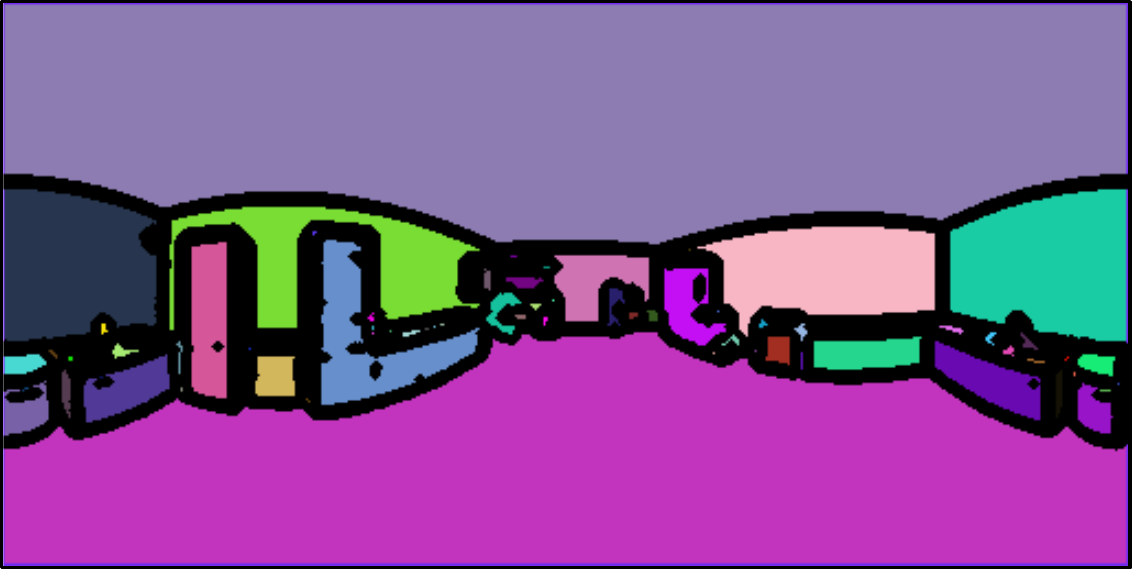}
\includegraphics[width=0.48\linewidth]{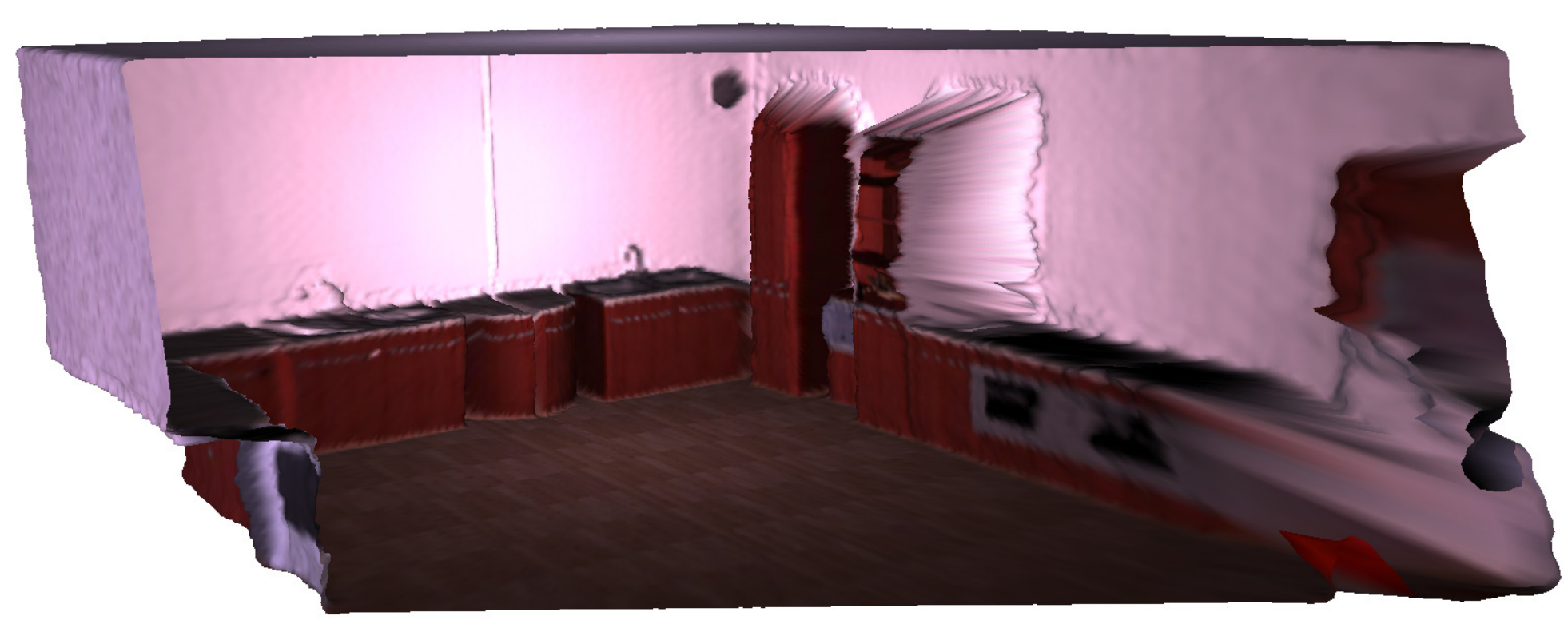}
\includegraphics[width=0.48\linewidth]{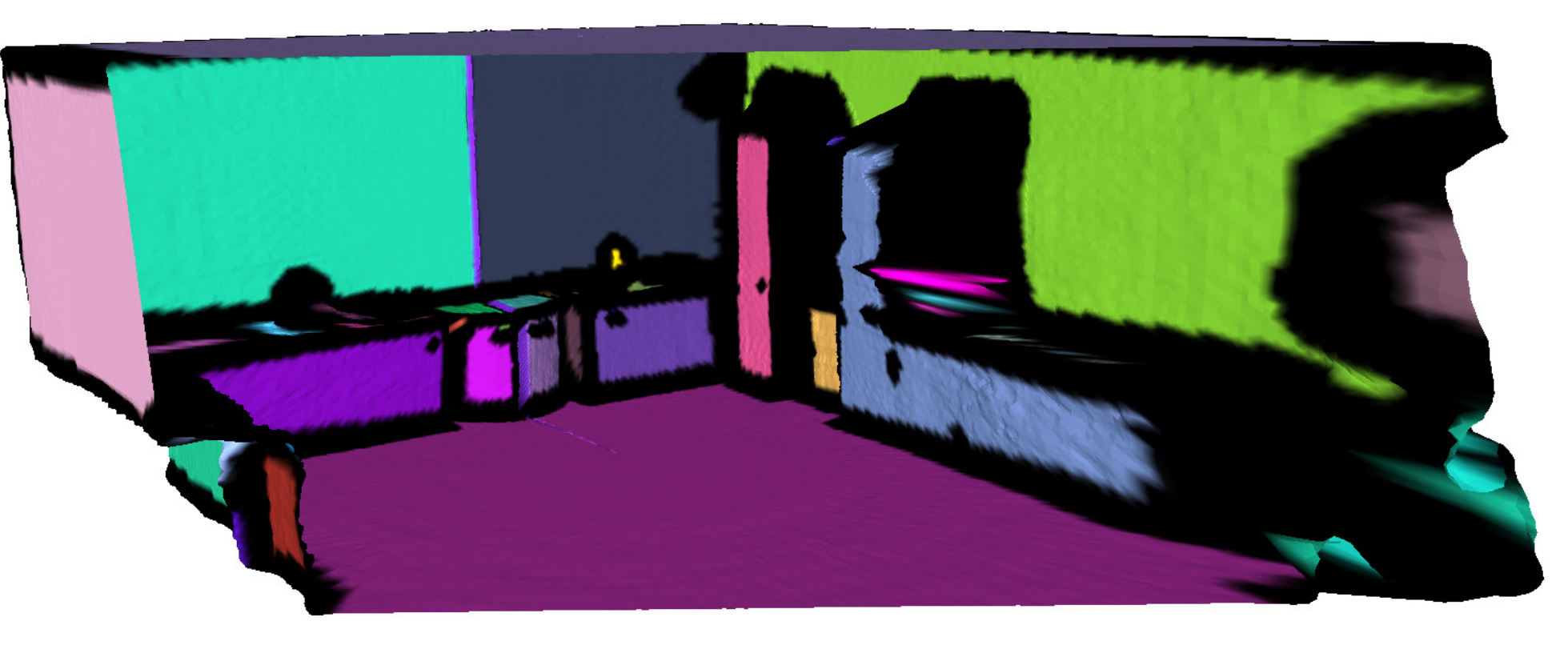}
\end{center}
\vspace{-2mm}
\caption{View of the 3D ``pop-up" model created from our network outputs. Left: our planar reconstruction textured with RGB image. Right: same model textured with plane segmentation.}
\label{fig:popup}
\end{figure}

To generate these reconstructions, we first isolate the sharpest edges in the planar boundary map using Otsu thresholding \cite{otsu1979threshold} and then identify each connected component in the resulting segmentation. An example of the result of this plane segmentation is shown in Figure \ref{fig:planeseg}. Thanks to the quality of our plane boundary predictions, this segmentation process requires no threshold tuning. To turn this segmentation into a 3D planar model, we first compute the median normal within each segmented plane. Then, we estimate the distance parameter of the plane equation in each segment using a 1-parameter RANSAC \cite{fischler1981random} with a final least-squares refinement over the inliers. Lastly, we project each pixel onto its associated plane. The model is finally ``popped-up" in 3D by back-projecting the point cloud according to these new depths. We mesh the points by resampling to the vertices of a icosahedral triangular grid and scaling the vertices according to the adjusted depths, resulting in the models shown in Figure \ref{fig:popup}.

\begin{figure}[ht]
\begin{center}
\includegraphics[width=0.48\linewidth]{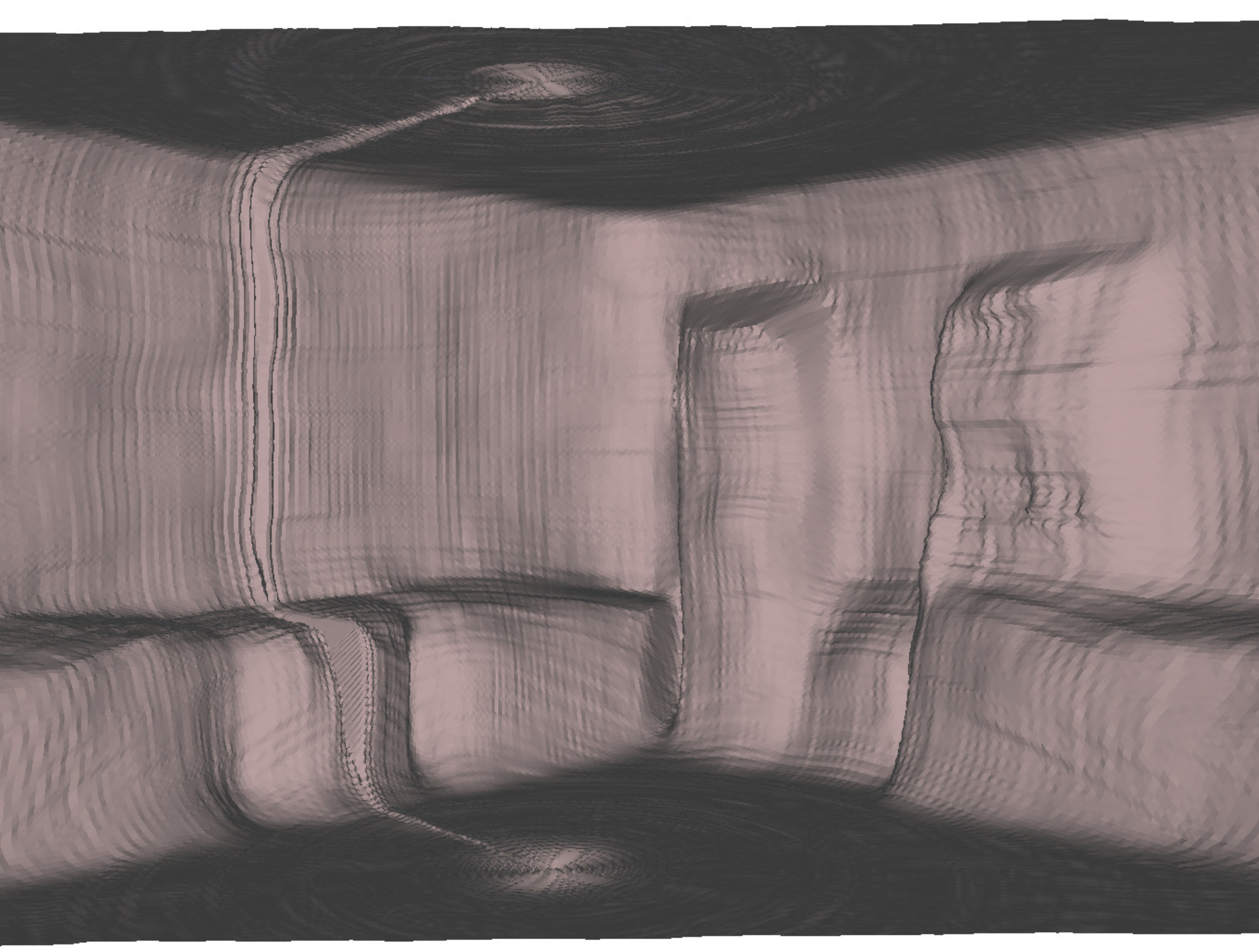}
\includegraphics[width=0.48\linewidth]{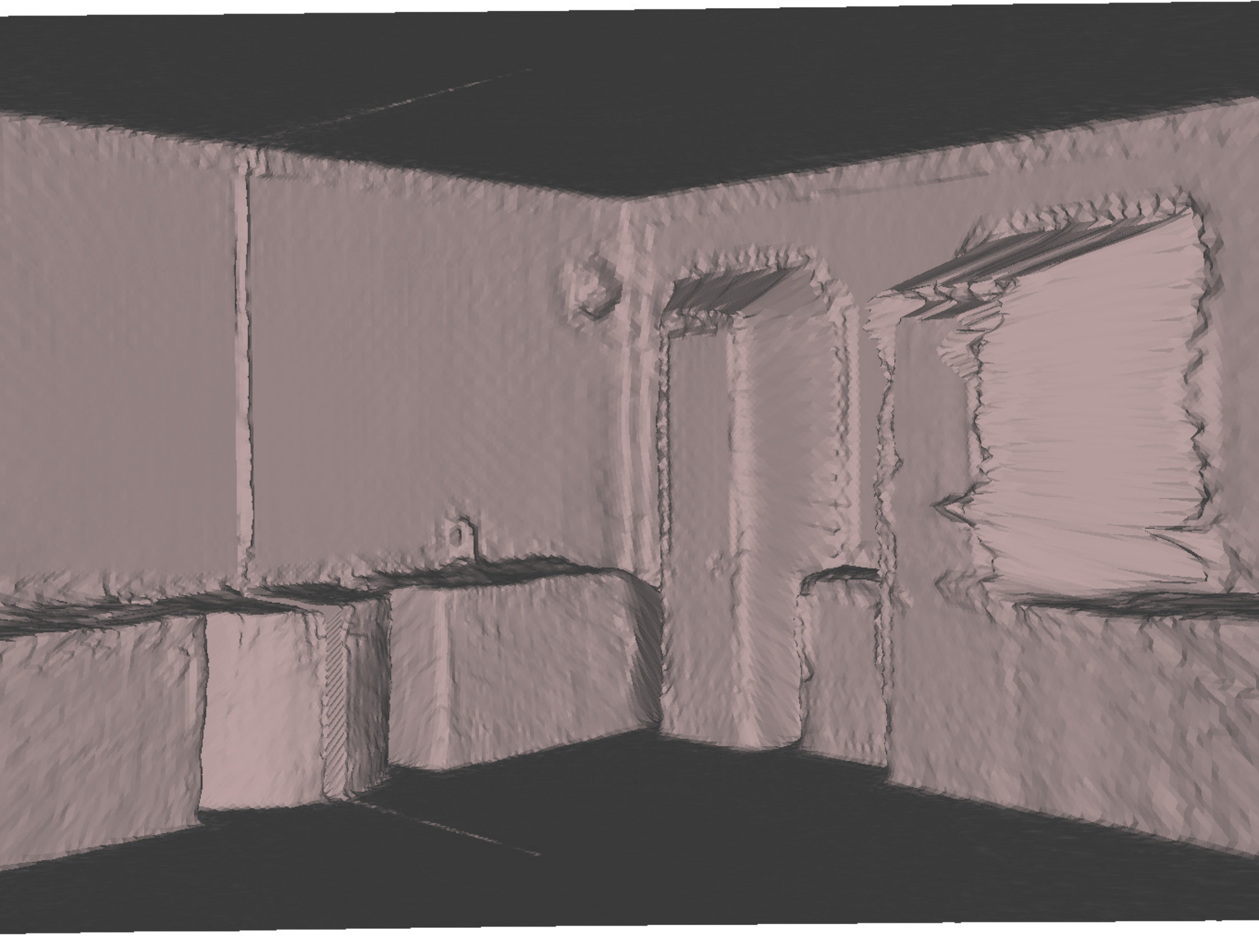}
\end{center}
\vspace{-2mm}
\caption{Left: Snapshot of an untextured, meshed 3D model produced from the baseline depth predictions using the image from Figure \ref{fig:popup}. Right: Equivalent popup model generated using the our proposed method.}
\label{fig:popupcomparison}
\end{figure}

Reiterating the importance of surface normal prediction, we found incorporating normal information to be vital to our RANSAC routine. Estimating planes solely from the depth estimates gives a much noisier reconstruction. Furthermore, we observe that having plane information allows us to produce higher quality 3D models than those generated from depth estimates alone. Figure \ref{fig:popupcomparison} compares our method, which leverages depth, normals, and boundary information, to the baseline network, which only estimates depth. Where the latter model suffers from smoothed edges, ours is able to produce sharp plane boundaries.

The significant drawback of monocular depth estimation is that the lack of any regularization over the estimates leads to fairly noisy predictions. This stands in contrast to stereo methods (and even pseudo-stereo methods like Godard \etal \cite{godard2017unsupervised}) in which a second image can be used to ensure consistency in the depth map. However, with our planar assumption, we can resolve some of the depth ambiguity while staying purely monocular. Moreover, the planar constraint removes the dependence on texture to recover depth.  Although making assumptions about the scene may be impractical for specific tasks like autonomous vehicle depth estimation \cite{smolyanskiy2018importance}, Figure \ref{fig:popupcomparison} demonstrates that a simple planarity assumption can be leveraged with great effect for indoor 3D modeling. 

\section{Conclusion}
We have presented a CNN capable of predicting depth, surface normals, and planar boundaries from a single indoor \threesixty image. Using a novel plane-aware loss function, we have achieved state-of-the-art results for these tasks. We have also demonstrated that the inclusion of a geodesic map can improve surface normal estimates for omnidrectional images. Lastly, we have shown that our network provides all the information necessary to produce a 3D planar model of the scene. Looking ahead, we see an emerging opportunity to utilize this type of all-in-one prediction from omnidirectional images to bootstrap indoor 3D reconstruction. 

{\small
\bibliographystyle{ieee}

}

\newpage
\section{Extended Results}
\begin{figure*}
    \centering
    \setlength{\fboxsep}{0pt}
    \setlength{\fboxrule}{0.5pt}

    \raisebox{23pt}{\parbox[b]{.03\textwidth}{1)}}%
    \begin{subfigure}[b]{0.23\textwidth}
        \centering
        \fbox{\includegraphics[width=\textwidth]{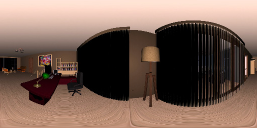}}
    \end{subfigure}
    ~
    \centering
    \begin{subfigure}[b]{0.23\textwidth}
        \centering
        \fbox{\includegraphics[width=\textwidth]{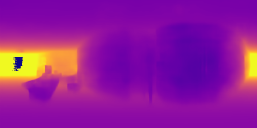}}
    \end{subfigure}
    ~
    \centering
    \begin{subfigure}[b]{0.23\textwidth}
        \centering
        \fbox{\includegraphics[width=\textwidth]{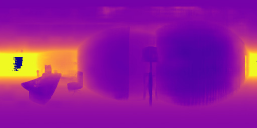}}
    \end{subfigure}
    ~
    \centering
    \begin{subfigure}[b]{0.23\textwidth}
        \centering
        \fbox{\includegraphics[width=\textwidth]{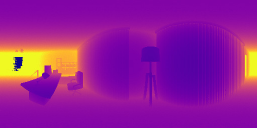}}
    \end{subfigure}\\
\vspace{2pt}

    \raisebox{23pt}{\parbox[b]{.03\textwidth}{2)}}%
    \begin{subfigure}[b]{0.23\textwidth}
        \centering
        \fbox{\includegraphics[width=\textwidth]{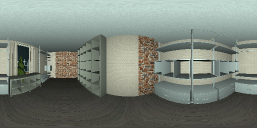}}
    \end{subfigure}
    ~
    \centering
    \begin{subfigure}[b]{0.23\textwidth}
        \centering
        \fbox{\includegraphics[width=\textwidth]{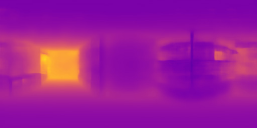}}
    \end{subfigure}
    ~
    \centering
    \begin{subfigure}[b]{0.23\textwidth}
        \centering
        \fbox{\includegraphics[width=\textwidth]{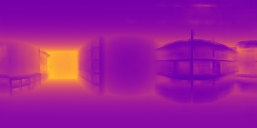}}
    \end{subfigure}
    ~
    \centering
    \begin{subfigure}[b]{0.23\textwidth}
        \centering
        \fbox{\includegraphics[width=\textwidth]{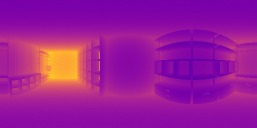}}
    \end{subfigure}\\
\vspace{2pt}
    
    \raisebox{23pt}{\parbox[b]{.03\textwidth}{3)}}%
    \begin{subfigure}[b]{0.23\textwidth}
        \centering
        \fbox{\includegraphics[width=\textwidth]{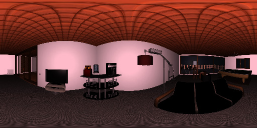}}
    \end{subfigure}
    ~
    \centering
    \begin{subfigure}[b]{0.23\textwidth}
        \centering
        \fbox{\includegraphics[width=\textwidth]{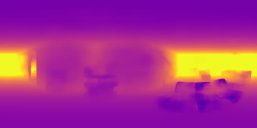}}
    \end{subfigure}
    ~
    \centering
    \begin{subfigure}[b]{0.23\textwidth}
        \centering
        \fbox{\includegraphics[width=\textwidth]{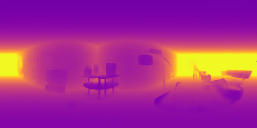}}
    \end{subfigure}
    ~
    \centering
    \begin{subfigure}[b]{0.23\textwidth}
        \centering
        \fbox{\includegraphics[width=\textwidth]{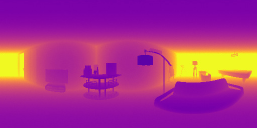}}
    \end{subfigure}\\
\vspace{2pt} 

    \raisebox{23pt}{\parbox[b]{.03\textwidth}{4)}}%
    \begin{subfigure}[b]{0.23\textwidth}
        \centering
        \fbox{\includegraphics[width=\textwidth]{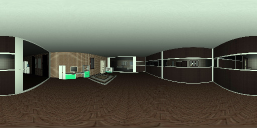}}
    \end{subfigure}
    ~
    \centering
    \begin{subfigure}[b]{0.23\textwidth}
        \centering
        \fbox{\includegraphics[width=\textwidth]{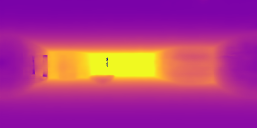}}
    \end{subfigure}
    ~
    \centering
    \begin{subfigure}[b]{0.23\textwidth}
        \centering
        \fbox{\includegraphics[width=\textwidth]{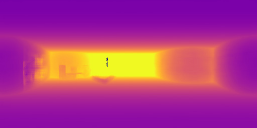}}
    \end{subfigure}
    ~
    \centering
    \begin{subfigure}[b]{0.23\textwidth}
        \centering
        \fbox{\includegraphics[width=\textwidth]{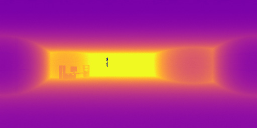}}
    \end{subfigure}\\
\vspace{2pt} 

    \raisebox{23pt}{\parbox[b]{.03\textwidth}{5)}}%
    \begin{subfigure}[b]{0.23\textwidth}
        \centering
        \fbox{\includegraphics[width=\textwidth]{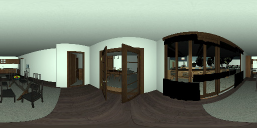}}
    \end{subfigure}
    ~
    \centering
    \begin{subfigure}[b]{0.23\textwidth}
        \centering
        \fbox{\includegraphics[width=\textwidth]{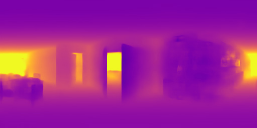}}
    \end{subfigure}
    ~
    \centering
    \begin{subfigure}[b]{0.23\textwidth}
        \centering
        \fbox{\includegraphics[width=\textwidth]{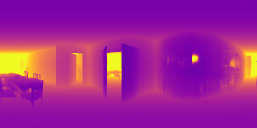}}
    \end{subfigure}
    ~
    \centering
    \begin{subfigure}[b]{0.23\textwidth}
        \centering
        \fbox{\includegraphics[width=\textwidth]{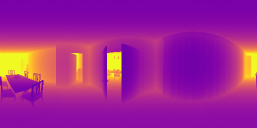}}
    \end{subfigure}\\
\vspace{2pt} 

    \raisebox{40pt}{\parbox[b]{.03\textwidth}{6)}}%
    \begin{subfigure}[b]{0.23\textwidth}
        \centering
        \fbox{\includegraphics[width=\textwidth]{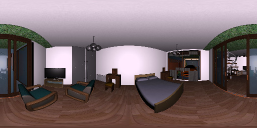}}
        \caption{RGB Input}
    \end{subfigure}
    ~
    \centering
    \begin{subfigure}[b]{0.23\textwidth}
        \centering
        \fbox{\includegraphics[width=\textwidth]{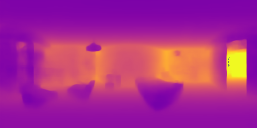}}
        \caption{Baseline}
    \end{subfigure}
    ~
    \centering
    \begin{subfigure}[b]{0.23\textwidth}
        \centering
        \fbox{\includegraphics[width=\textwidth]{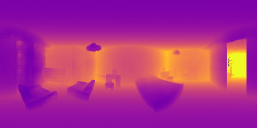}}
        \caption{Ours}
    \end{subfigure}
    ~
    \centering
    \begin{subfigure}[b]{0.23\textwidth}
        \centering
        \fbox{\includegraphics[width=\textwidth]{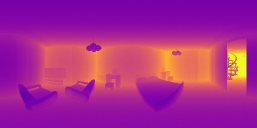}}
        \caption{Ground Truth}
    \end{subfigure}\\
\caption{More qualitative comparisons of depth predictions using our plane-aware loss compared to the baseline method based on Zioulis \etal \cite{zioulis2018omnidepth}. Our results are noticeably better at capturing the depth of planar objects in the scenes, such as the shelves in row (2) or the table in row (5). Row (3) shows a case where our method in unable to capture a large planar section, but it is worth noting that the baseline method was unsuccessful as well.}
\label{fig:suppqualitativedepth}
\end{figure*}

\begin{figure*}
    \centering
    \setlength{\fboxsep}{0pt}
    \setlength{\fboxrule}{0.5pt}

\vspace{2pt}    
    \raisebox{23pt}{\parbox[b]{.03\textwidth}{1)}}%
    \begin{subfigure}[b]{0.23\textwidth}
        \centering
        \fbox{\includegraphics[width=\textwidth]{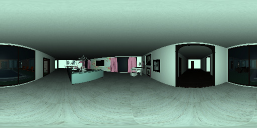}}
    \end{subfigure}
    ~
    \centering
    \begin{subfigure}[b]{0.23\textwidth}
        \centering
        \fbox{\includegraphics[width=\textwidth]{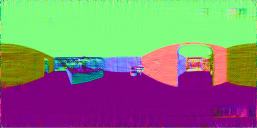}}
    \end{subfigure}
    ~
    \centering
    \begin{subfigure}[b]{0.23\textwidth}
        \centering
        \fbox{\includegraphics[width=\textwidth]{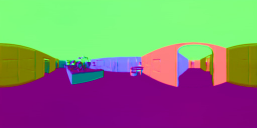}}
    \end{subfigure}
    ~
    \centering
    \begin{subfigure}[b]{0.23\textwidth}
        \centering
        \fbox{\includegraphics[width=\textwidth]{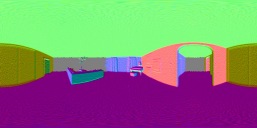}}
    \end{subfigure}\\

\vspace{2pt}    
    \raisebox{23pt}{\parbox[b]{.03\textwidth}{2)}}%
    \begin{subfigure}[b]{0.23\textwidth}
        \centering
        \fbox{\includegraphics[width=\textwidth]{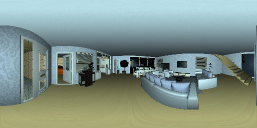}}
    \end{subfigure}
    ~
    \centering
    \begin{subfigure}[b]{0.23\textwidth}
        \centering
        \fbox{\includegraphics[width=\textwidth]{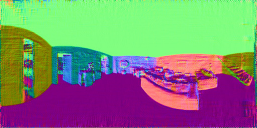}}
    \end{subfigure}
    ~
    \centering
    \begin{subfigure}[b]{0.23\textwidth}
        \centering
        \fbox{\includegraphics[width=\textwidth]{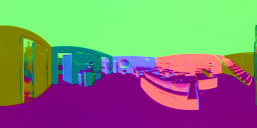}}
    \end{subfigure}
    ~
    \centering
    \begin{subfigure}[b]{0.23\textwidth}
        \centering
        \fbox{\includegraphics[width=\textwidth]{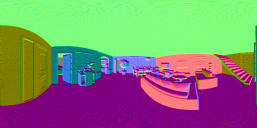}}
    \end{subfigure}\\

    \raisebox{23pt}{\parbox[b]{.03\textwidth}{3)}}%
    \begin{subfigure}[b]{0.23\textwidth}
        \centering
        \fbox{\includegraphics[width=\textwidth]{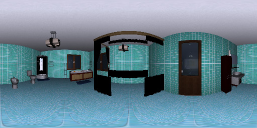}}
    \end{subfigure}
    ~
    \centering
    \begin{subfigure}[b]{0.23\textwidth}
        \centering
        \fbox{\includegraphics[width=\textwidth]{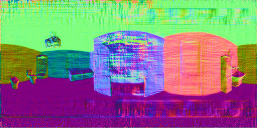}}
    \end{subfigure}
    ~
    \centering
    \begin{subfigure}[b]{0.23\textwidth}
        \centering
        \fbox{\includegraphics[width=\textwidth]{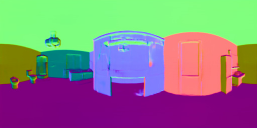}}
    \end{subfigure}
    ~
    \centering
    \begin{subfigure}[b]{0.23\textwidth}
        \centering
        \fbox{\includegraphics[width=\textwidth]{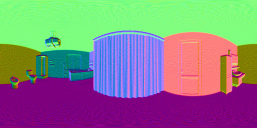}}
    \end{subfigure}\\
\vspace{2pt}
    
    \raisebox{23pt}{\parbox[b]{.03\textwidth}{4)}}%
    \begin{subfigure}[b]{0.23\textwidth}
        \centering
        \fbox{\includegraphics[width=\textwidth]{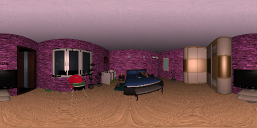}}
    \end{subfigure}
    ~
    \centering
    \begin{subfigure}[b]{0.23\textwidth}
        \centering
        \fbox{\includegraphics[width=\textwidth]{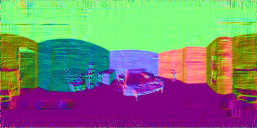}}
    \end{subfigure}
    ~
    \centering
    \begin{subfigure}[b]{0.23\textwidth}
        \centering
        \fbox{\includegraphics[width=\textwidth]{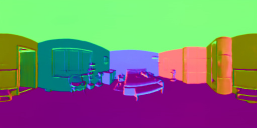}}
    \end{subfigure}
    ~
    \centering
    \begin{subfigure}[b]{0.23\textwidth}
        \centering
        \fbox{\includegraphics[width=\textwidth]{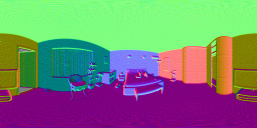}}
    \end{subfigure}\\
\vspace{2pt}

    \raisebox{23pt}{\parbox[b]{.03\textwidth}{5)}}%
    \begin{subfigure}[b]{0.23\textwidth}
        \centering
        \fbox{\includegraphics[width=\textwidth]{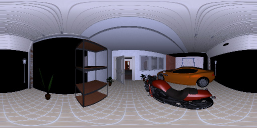}}
    \end{subfigure}
    ~
    \centering
    \begin{subfigure}[b]{0.23\textwidth}
        \centering
        \fbox{\includegraphics[width=\textwidth]{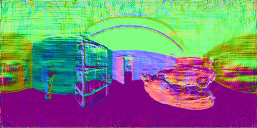}}
    \end{subfigure}
    ~
    \centering
    \begin{subfigure}[b]{0.23\textwidth}
        \centering
        \fbox{\includegraphics[width=\textwidth]{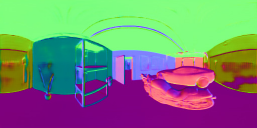}}
    \end{subfigure}
    ~
    \centering
    \begin{subfigure}[b]{0.23\textwidth}
        \centering
        \fbox{\includegraphics[width=\textwidth]{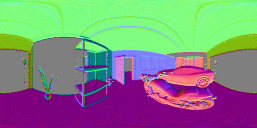}}
    \end{subfigure}\\

\vspace{2pt}    
    \raisebox{40pt}{\parbox[b]{.03\textwidth}{6)}}%
    \begin{subfigure}[b]{0.23\textwidth}
        \centering
        \fbox{\includegraphics[width=\textwidth]{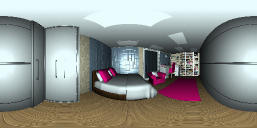}}
        \caption{RGB Input}
    \end{subfigure}
    ~
    \centering
    \begin{subfigure}[b]{0.23\textwidth}
        \centering
        \fbox{\includegraphics[width=\textwidth]{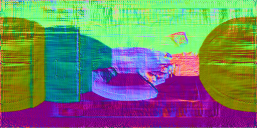}}
        \caption{Derived}
    \end{subfigure}
    ~
    \centering
    \begin{subfigure}[b]{0.23\textwidth}
        \centering
        \fbox{\includegraphics[width=\textwidth]{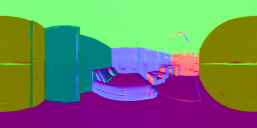}}
        \caption{Predicted}
    \end{subfigure}
    ~
    \centering
    \begin{subfigure}[b]{0.23\textwidth}
        \centering
        \fbox{\includegraphics[width=\textwidth]{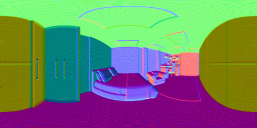}}
        \caption{Ground Truth}
    \end{subfigure}\\
\vspace{-2mm}
\caption{Extended comparison between our surface normal predictions and those derived the from depth predictions alone. Rows (1) and (2) gives more examples where the normals derived from depth perform well, but rows (3)-(6) show that, generally, we are better off predicting normals independently from depth.}
\label{fig:suppnormalresults}
\end{figure*}

\begin{figure*}
    \centering
    \setlength{\fboxsep}{0pt}
    \setlength{\fboxrule}{0.5pt}

    \begin{subfigure}[b]{0.3\textwidth}
        \centering
        \fbox{\includegraphics[width=\textwidth]{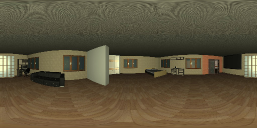}}
    \end{subfigure}
    ~
    \centering
    \begin{subfigure}[b]{0.3\textwidth}
        \centering
        \fbox{\includegraphics[width=\textwidth]{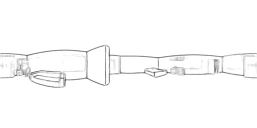}}
    \end{subfigure}
    ~
    \centering
    \begin{subfigure}[b]{0.3\textwidth}
        \centering
        \fbox{\includegraphics[width=\textwidth]{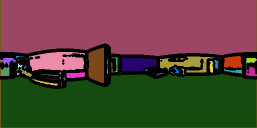}}
    \end{subfigure}\\
    ~
    \centering
    \begin{subfigure}[b]{0.7\textwidth}
        \centering
        \fbox{\includegraphics[width=\textwidth]{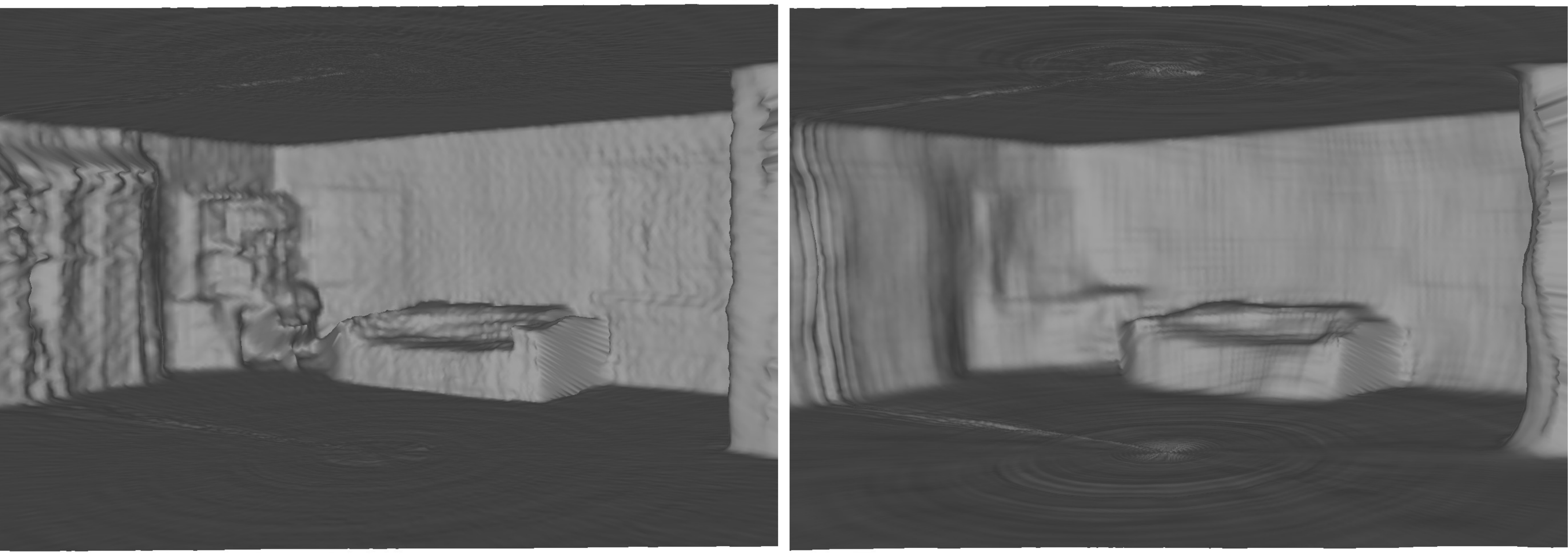}}
    \end{subfigure}\\
\vspace{2pt}

    \begin{subfigure}[b]{0.3\textwidth}
        \centering
        \fbox{\includegraphics[width=\textwidth]{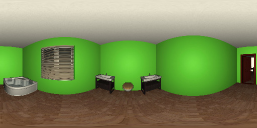}}
    \end{subfigure}
    ~
    \centering
    \begin{subfigure}[b]{0.3\textwidth}
        \centering
        \fbox{\includegraphics[width=\textwidth]{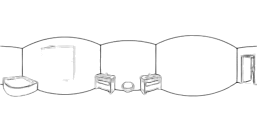}}
    \end{subfigure}
    ~
    \centering
    \begin{subfigure}[b]{0.3\textwidth}
        \centering
        \fbox{\includegraphics[width=\textwidth]{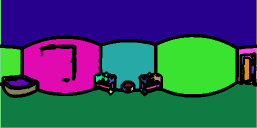}}
    \end{subfigure}\\
    ~
    \centering
    \begin{subfigure}[b]{0.7\textwidth}
        \centering
        \fbox{\includegraphics[width=\textwidth]{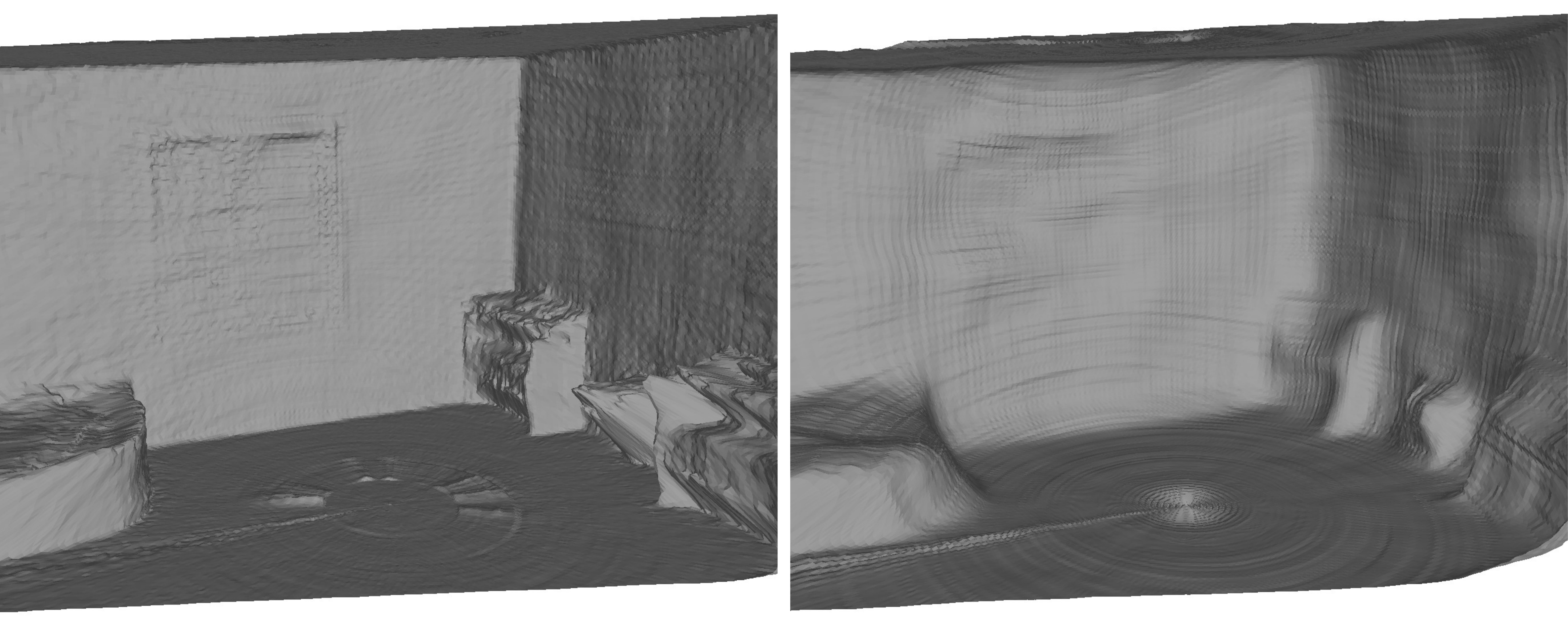}}
    \end{subfigure}\\
\vspace{2pt}

\caption{More comparisons of plane segmentations and 3D popups. For each example, the top row shows RGB input, plane boundary prediction, and plane segmentation, respectively, from left to right. Beneath those are a comparison of our popup reconstruction (left) and a mesh constructed from the baseline depth estimation (right). We show the untextured mesh to better highlight the differences in geometry. The rough regions in our reconstruction fall on the boundaries of the plane segmentation, highlighting that our method, while generally useful, falls prey to `fat edges' on the plane boundaries.}
\label{fig:supppopupcomp}
\end{figure*}

In this section, we provide a further qualitative review of our work. Figure \ref{fig:suppqualitativedepth} shows more examples of our network's depth estimates compared to our baseline. Figure \ref{fig:suppnormalresults} provides more cases to justify the prediction of normals independently from depth. Finally, Figure \ref{fig:supppopupcomp} shows more comparisons of popup reconstructions along with examples of the plane boundary predictions and segmentations.

\vspace{-1mm}

\end{document}